\documentclass[10pt,twocolumn,letterpaper]{article}

 \usepackage[pagenumbers]{cvpr} 

\usepackage{algorithm}
\usepackage{algorithmic}

\definecolor{cvprblue}{rgb}{0.21,0.49,0.74}
\usepackage[pagebackref,breaklinks,colorlinks,allcolors=cvprblue]{hyperref}

\def\confName{CVPR}
\def\confYear{2027}

\title{AG-WAM: Attention-Guided World Action Model for Accurate and Efficient Robot Manipulation}

\author{
   {Hongcheng Zhang}$^\ast$ \quad
   {Lei Sun}$^\ast$ \quad
   {Songxuan Liu}$^\ast$ \quad
   {Tengteng Pan} \\
   {Yuqiang Zhai} \quad
   {Xianglong Xu} \quad
   {Jianya Guo} \quad
   {Xiao Song}$^\dagger$ \\
   {SenseTime Research} \\
   {\tt\small \{zhanghongcheng,sunlei,liusongxuan,songxiao\}@sensetime.com }
}

\newcommand\nnfootnote[1]{
  \begin{NoHyper}
  \renewcommand\thefootnote{}\footnote{#1}
  \addtocounter{footnote}{-1}
  \end{NoHyper}
}

\begin{document}
\maketitle

\nnfootnote{ $^\ast$ Equal Contribution. $^\dagger$ Corresponding Author.}

\begin{abstract}
Recently developed World Action Models (WAMs) jointly predict future observations and robot actions, providing predictive context for control. However, existing WAMs typically learn implicit attention allocation with limited explicit action-relevant attention guidance, leading to limited performance. To unleash strong capabilities of WAMs, we introduce explicit attention guidance as a unifying principle and propose our Attention-Guided WAM (AG-WAM), an end-to-end framework with three plug-and-play attention guidance modules that directly reformulate internal attention paradigm in WAMs: the Content-Adaptive Predictive Attention module, the Guidance-Point Visual Attention module, and the State-Conditioned Text Attention module. Without any external plugin module adopted, our end-to-end framework with the unifying principle of explicit attention guidance boosts the accuracy and efficiency of WAMs. Without using any private embodied training data, our AG-WAM establishes new state-of-the-art performance on multiple public robotic manipulation benchmarks including LIBERO-X, LIBERO-Plus, and RoboTwin-2.0, significantly outperforming  other methods, even those using massive private embodied training data by a notable margin, meanwhile notably improving training and inference efficiency.
\end{abstract}
\section{Introduction}
\label{sec:intro}

World models for robotics aim to predict how the physical world evolves in response to an agent’s actions. Building on this foundation, World Action Models~\cite{wang2026wam,zhu2025uwm,ye2026dreamzero} jointly model future visual observations and robot actions to support precise and robust robotic control. Compared with policies that directly map observations and task instructions to actions, WAMs incorporate an additional visual prediction objective to capture dynamics and temporal dependencies relevant to control, providing predictive context for action generation. However, achieving accurate control meanwhile maintaining inference efficiency remains challenging.


\begin{figure}
\centering
\includegraphics[width=0.5\textwidth]{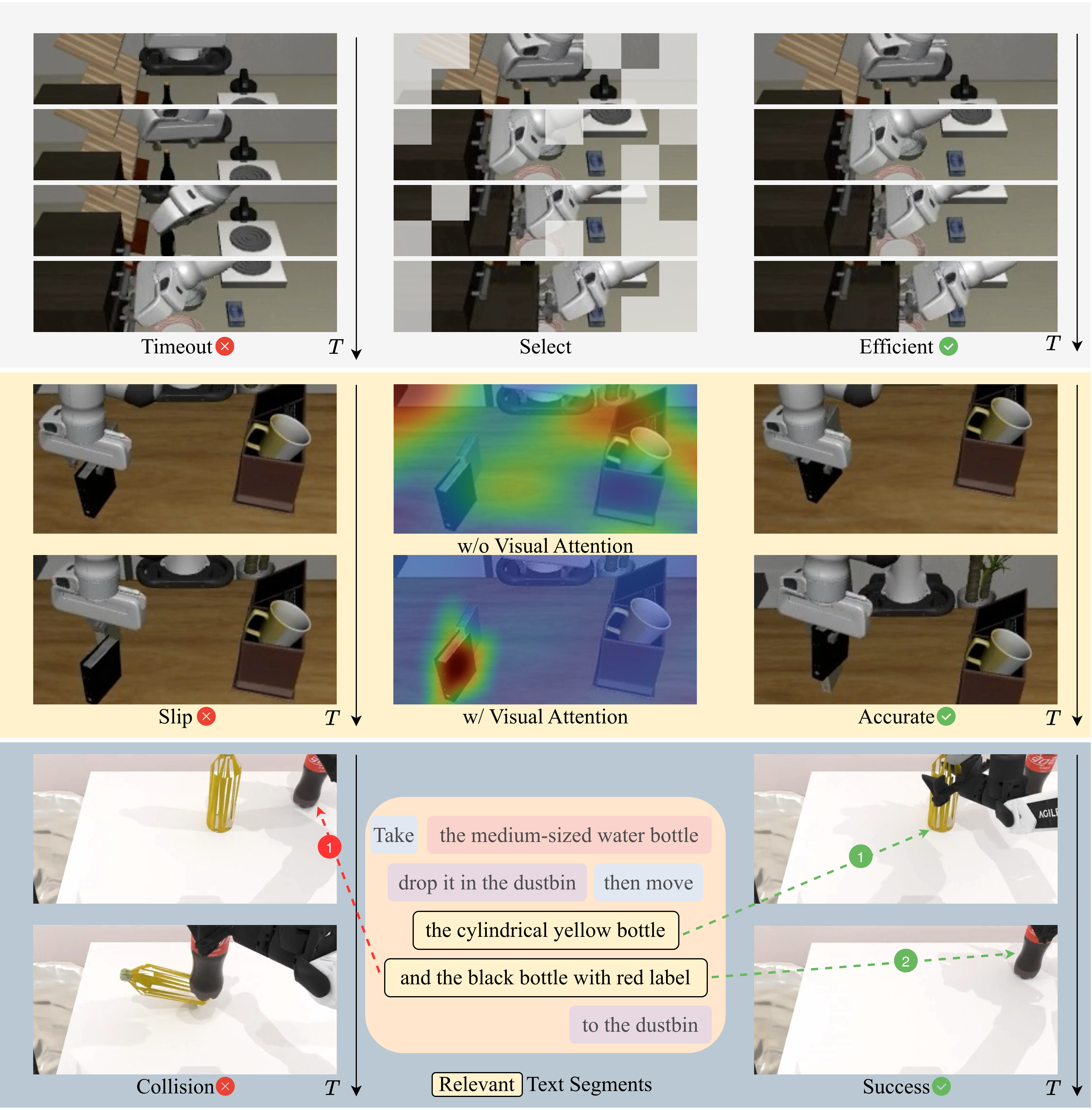}
\caption{Overview examples of our proposed explicit attention guidance principle. Top: attention selectively allocated to task-relevant video tokens that are retained for subsequent processing, accelerating inference meanwhile maintaining accuracy. Middle: visual attention explicitly directed toward interaction-relevant regions (black book), enabling accurate manipulation. Bottom: instruction attention explicitly guided toward relevant text segments based on the evolving execution state.}
\label{fig:Fig1}
\vspace{-0.3cm}
\end{figure}

Attention plays a central role in WAMs by determining how  action stream accesses visual observations and task instructions. In existing WAMs~\cite{ye2026dreamzero,bi2026motus,li2026lingbotva,yuan2026fastwam}, however, attention allocation is typically learned implicitly, and action stream may attend to visual regions or text tokens with limited relevance to the current execution stage. This observation motivates us to introduce explicit attention guidance toward action-relevant visual and language context.

To this end, we propose Attention-Guided WAM, an end-to-end framework that incorporates explicit attention guidance  for action generation. AG-WAM provides unified attention guidance centered on action-relevant information by selectively allocating attention to informative video tokens, directing visual attention toward interaction-relevant regions, and guiding instruction attention according to the evolving execution state. Accordingly, we propose three plug-and-play explicit attention guidance modules.

The Content-Adaptive Predictive Attention (CAPA) module uses spatiotemporal information from the observed video history to predict video token retention indices for the next chunk. By restricting subsequent video token processing to the retained subset, CAPA allocates attention to task-relevant information meanwhile reducing computational burden. The Guidance-Point Visual Attention (GPVA) module infers guidance points from visual observation, task context, and robot proprioception. GPVA incorporates complementary 2D appearance and 3D geometric priors into visual attention, guiding action tokens to attend to interaction-relevant visual regions. The State-Conditioned Text Attention (SCTA) module explicitly guides instruction attention toward relevant text segments based on the evolving execution state, in which a state-dependent semantic condition is first produced, then used to integrate relevant instruction information into action features through feature-wise linear modulation (FiLM)~\cite{perez2018film}. Overall, built on the unifying principle of explicit attention guidance, our proposed AG-WAM integrates the CAPA, GPVA, and SCTA modules, providing complementary attention guidance that explicitly directs attention toward action-relevant information, unleashing strong capabilities of WAMs, and ultimately achieving accurate and efficient action generation.
We evaluate AG-WAM on three public robotic manipulation benchmarks: LIBERO-X~\cite{wang2026liberox}, LIBERO-Plus~\cite{fei2025liberoplus}, and RoboTwin-2.0~\cite{chen2025robotwin2}. Without using any private embodied training data, AG-WAM achieves state-of-the-art performance on all three benchmarks with success rates of $\mathbf{50.2\%}$, $\mathbf{95.1\%}$, and $\mathbf{97.14\%}$ respectively, outperforming other methods by a notable margin, meanwhile delivering a $1.45\times$ inference speedup. These results  demonstrate the effectiveness of the unifying principle of explicit attention guidance. 
Our contributions are as follows:
\begin{enumerate}
    \item We address the problem of implicit attention allocation in WAMs. To this end, we introduce explicit attention guidance as a unifying principle to unleash strong capabilities of WAMs. Formally, we propose AG-WAM, an end-to-end framework for accurate and efficient action generation, without any external plugin module adopted.

    
    \item We propose three plug-and-play attention guidance modules that directly reformulate internal attention paradigm in WAMs: CAPA predicts token retention guiding attention to select video tokens most informative for action generation; GPVA directs visual attention toward interaction-relevant regions using 2D appearance and 3D geometric priors; SCTA guides instruction attention based on the evolving execution state. 
    

    \item AG-WAM establishes new state-of-the-art performance on three public benchmarks, significantly outperforming other methods, even those using massive private embodied training data by a notable margin, meanwhile notably improving training and inference efficiency. 
    
\end{enumerate}
\section{Related Work}
\label{sec:related_work}

\subsection{World Action Models}
\label{sec:related_wam}

\begin{figure*}
\centering
\includegraphics[width=0.9\textwidth]{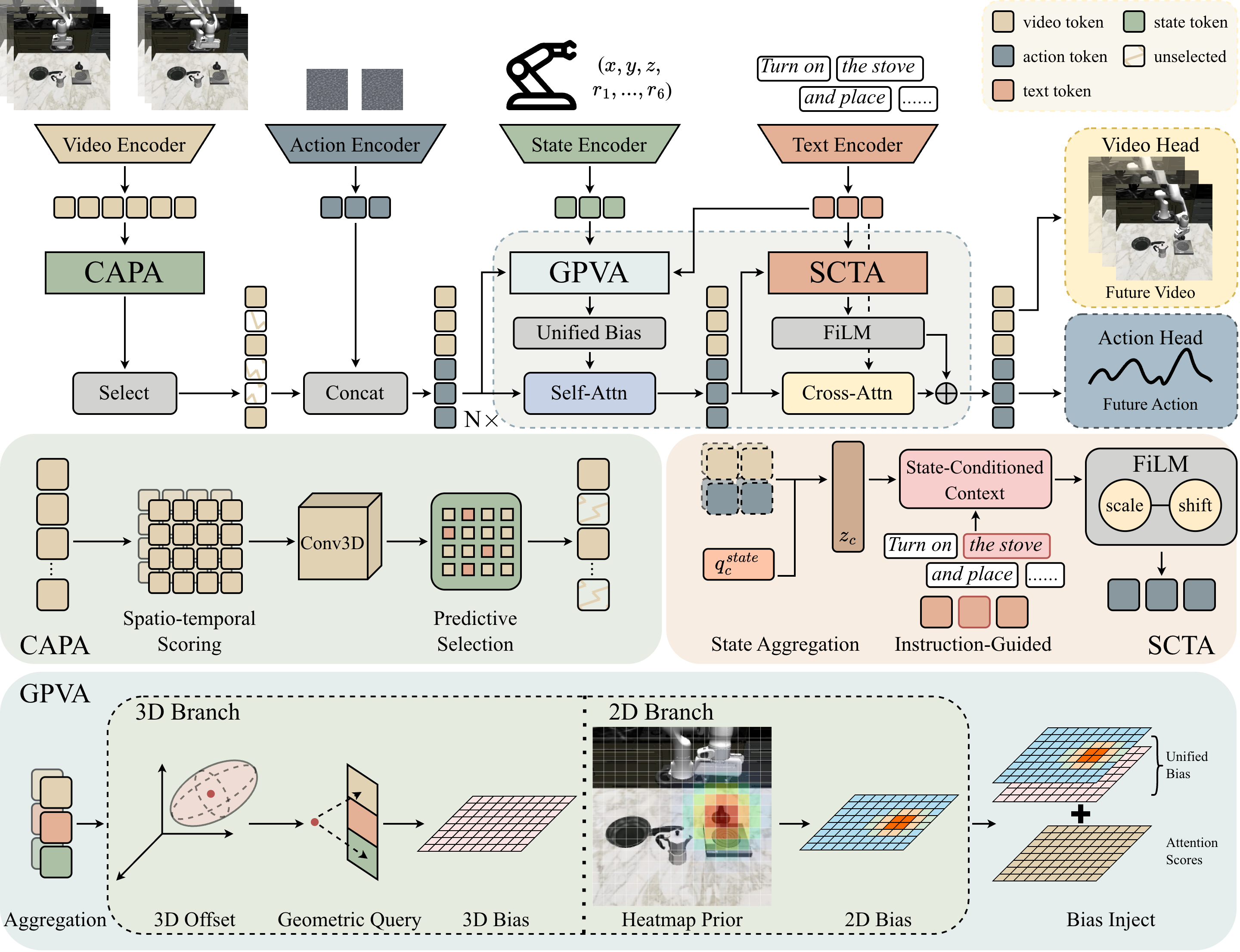}
\vspace{-0.2cm}
\caption{The overall architecture of AG-WAM, an end-to-end framework integrating three explicit attention guidance modules.}
\label{fig:Fig2}
\end{figure*}

Different from Vision‑Language‑Action (VLA) policies that map observations to actions~\cite{brohan2023rt2, kim2025openvlaoft, team2024octo, black2024pi0}, WAMs jointly model forward visual dynamics and action generation to capture physical foresight~\cite{wang2026wam}. Existing architectures generally follow two paradigms: Cascade WAMs first synthesize future frames~\cite{du2023unipi, bharadhwaj2024gen2act}, intermediate diffusion features~\cite{hu2024vpp}, or visual goals~\cite{pi2026pi07} before decoding actions; Joint WAMs co-optimize video prediction and action generation within a shared network via autoregression~\cite{wu2024gr1, cheang2024gr2, cen2025worldvla}, predictive inverse dynamics~\cite{tian2025seer}, unified diffusion or flow-matching~\cite{guo2024pad, li2025uva, zhu2025uwm, bi2026motus, xiang2026motubrain, li2026lingbotva, zhang2026lingbotva2},  and representation alignment~\cite{qiu2026agra}. Recent works augment robot manipulation pretraining with web-scale egocentric videos to model physical dynamics and latent actions~\cite{wang2023mimicplay, gao2026dreamdojo, ye2025lapa}. However, corresponding representations may entangle with complex generalizations of multi-modal context. 

To improve the capability or efficiency of WAMs, recent works mainly pursue through external avenues: 
attaching external multi-modal reasoners~\cite{zhao2025cotvla, tang2026adawam, lin2026worldpilot}, which correspondingly introduces the high latency and distributional mismatch with low-level control; or bypassing the test-time video generation~\cite{li2025uva, yuan2026fastwam, ma2026fasterwam, ye2026gigaworld}. 

\subsection{Attention Guidance}
\label{sec:related_attention}

\noindent\textbf{Token Selection.}
To reduce redundancy, a few WAMs rely on fixed-stride temporal downsampling~\cite{bi2026motus, ye2026gigaworld} that ignores spatial foreground-background differences. Meanwhile, token pruning and merging in VLA~\cite{chen2024fastv, xu2026vlacache, yang2026efficientvla} and video diffusion~\cite{bolya2023tomesd, xi2025sparsevideogen, xia2025adaspa} are agnostic to future action generation, 
limiting access to task-relevant information.

\smallskip\noindent\textbf{Visual Guidance.}
Prior manipulation policies inject spatial cues by querying external VLMs for 2D keypoints or constraints~\cite{yuan2024robopoint, fang2024moka, huang2024rekep}, overlaying visual markers onto RGB inputs~\cite{nasiriany2025rtaffordance, zheng2025tracevla}, or adding static 3D encodings~\cite{qu2025spatialvla}. However, these methods rely on slow external two-stage VLMs or extra input augmentations.

\smallskip\noindent\textbf{Instruction Conditioning.}
Action stream accesses multi-modal context either indiscriminately~\cite{pi2026pi07, hu2024vpp} or through predefined attention masks~\cite{liang2024mot, zhu2025uwm, bi2026motus, li2026lingbotva, yuan2026fastwam}. 
However, these methods lack explicit guidance to adapt instruction attention to the evolving execution state.


\section{Method}
\label{sec:method}

\subsection{Framework}
\label{subsec:Framework}
Building on prior WAM architectures \cite{li2026lingbotva,xiang2026motubrain}, we develop AG-WAM upon Wan2.2-5B \cite{wan2025wan}.
RGB observations are encoded into latents by the VAE and converted into video tokens through patch embedding. Task instructions are encoded by a frozen T5 encoder and projected into text embeddings $E_{text}$, and an MLP maps quantile-normalized actions to action tokens. During training, clean video and action tokens provide context, while noisy video and action tokens are denoised through flow matching. These tokens are interleaved in temporal order and concatenated as a unified input $E_{v,a}$. 
Under a structured causal mask, the backbone processes $E_{v,a}$ with $E_{text}$ injected through cross-attention:
\begin{equation}
\bigl(\hat F^{v},\hat F^{a}\bigr)
=f_\theta\!\left(E_{v,a},\tau_v,\tau_a;E_{text}\right)
\label{eq:framework}
\end{equation}
where $\tau_v$ and $\tau_a$ denote video and action noise levels, $\hat F^{v}$ and $\hat F^{a}$ denote corresponding flow velocity predictions. 
Built upon the proposed baseline framework above, AG-WAM is established by integrating three attention guidance modules. In Fig.~\ref{fig:Fig2}, CAPA is placed before the transformer backbone, predicting token retention guiding attention to select video tokens informative for action generation. GPVA and SCTA are inserted in transformer backbone with shared parameters across layers, grounding action-to-visual and action-to-instruction attention for action modulation. 

\subsection{Content-Adaptive Predictive Attention}
\label{sec:capa}

Video tokens contain massive redundant background information. Thus selectively retaining task-relevant tokens is essential. Accordingly, we propose the Content-Adaptive Predictive Attention module, which predicts video token retention indices based on spatiotemporal representations from the observed video history, allocating task-relevant attention meanwhile reducing computational burden.


\noindent\textbf{Spatiotemporal Scoring for Predictive Selection.}
For the observed chunk $c$, after VAE encoding, patch embedding is adopted to partition each frame into an $H_g\times W_g$ grid with a $D$-dim projected vector per patch. We denote the resulting clean video token embeddings by $E_c\in\mathbb{R}^{T_c\times S\times D}$, where $T_c$ denotes the number of frames and $S=H_g\times W_g$ denotes the number of spatial tokens at each frame.


We project $E_c$ from dimension $D$ to $d_r$, yielding $X_c$. Let $x_{c,t,s}$ denote the feature in $X_c$ at the frame $t$ and the spatial position $s$, and $\bar{x}_{c,s}$ the temporal mean feature within the chunk $c$. We construct the input of the scoring network:
\begin{equation}
u_{c,t,s}
=\operatorname{Concat}\!\left(
x_{c,t,s},\,\bar{x}_{c,s},\,x_{c,t,s}-x_{c-1,t,s}
\right)
\label{eq:capa-context}
\end{equation}
The scoring network is designed as a depthwise-separable 3D convolution block, followed by an MLP head and the sigmoid activation. $u_{c,t,s}$ is rearranged into the ordered spatiotemporal feature $U_c$, which is then fed to the scoring network, predicting a retention score for each spatial token in the chunk $c+1$ based on the observed content of chunk $c$.


\noindent\textbf{Content-Guided Sparse Attention.}
We combine predictive selection with random sampling to determine the retention indices. Each frame retains $N_{\mathrm{ret}}=\lfloor\eta_{\mathrm{ret}}S\rfloor$ spatial tokens where $\eta_{\mathrm{ret}}$ denotes the  retention ratio: 
\begin{equation}
\mathcal I_{c+1,t}=\mathcal I^s_{c+1,t}\cup\mathcal I^r_{c+1,t}
\label{eq:capa-selection}
\end{equation}
where $\mathcal I^s_{c+1,t}$ contains indices with the highest scores, and $\mathcal I^r_{c+1,t}$ is sampled randomly from the remaining indices, additionally introducing robustness into token selection. Hence $N_{\mathrm{ret}}$ spatial tokens are exactly retained at each frame. 
Let $E_v$ 
denote video 
token embeddings of the chunk $c+1$.
We collect all retained subsets across frames as $\mathcal I=\{\mathcal I_{c+1,t}\}_{t=1}^{T_{c+1}}$ and construct the retained video token sequence as $\widetilde E_v=\operatorname{Gather}(E_v,\mathcal I)$.


The same retention indices are adopted to gather the corresponding positional encoding and metadata, 
concentrating computation on retained video tokens meanwhile preserving the original spatiotemporal structure and maintaining the causal attention constraint across all layers.

\noindent\textbf{Training and Inference.} Since discrete retention indices make the token selection operation non-differentiable, gradients cannot flow directly to the scoring network. We therefore train the scoring network using the surrogate-gradient~\cite{bengio2013estimating} update. Let $h^v_{\ell,i}$ and $\hat h^v_{\ell,i}$ denote input and output hidden states of the $i$-th retained video token at the layer $\ell$, $g_i$ denote the corresponding surrogate weight, and $\operatorname{sg}$ denote the stop-gradient operator:
\begin{equation}
h^v_{\ell+1,i}
=
\hat h^v_{\ell,i}
+
\left(g_i-\operatorname{sg}(g_i)\right)
\operatorname{sg}\!\left(\hat h^v_{\ell,i}-h^v_{\ell,i}\right)
\label{eq:capa-surrogate}
\end{equation}
The correction term
$\left(g_i-\operatorname{sg}(g_i)\right)
\operatorname{sg}(\hat h^v_{\ell,i}-h^v_{\ell,i})$
is zero in the forward propagation but supplies an approximate gradient to the scoring network during backpropagation. 
The transformer backbone and scoring network are jointly trained with the video-action loss, computed over retained video tokens and all action tokens.
\emph{More details are provided in the Supplementary Material.}

At inference, CAPA determines the retained subset for the chunk $c+1$ from the observed context in the chunk $c$, which remains unchanged across transformer layers and denoising steps. Only retained video tokens are denoised to provide future visual context for action generation, without reconstructing complete video frames. During video denoising, each layer caches the K/V of retained video tokens, and the subsequent action denoising reuses the K/V from the final video denoising step. Overall, processing fewer video tokens notably reduces computational burden.
\subsection{Guidance-Point Visual Attention}
\label{sec:GPVA}

Implicitly learned visual attention allocation typically incurs dispersive attention distribution and erroneous attendance to distracting objects or irrelevant regions. To this end, we propose the Guidance-Point Visual Attention module, in which complementary 2D appearance and 3D geometric priors are incorporated, guiding visual attention toward interaction-relevant regions and aggregating informative spatial context for accurate action generation.

\noindent\textbf{Guidance Point Formulation.} 
We represent the current goal of robot as the guidance point, such as a grasp or placement position. For a future frame $t$ in the target action chunk,  $p_t\in\mathbb{R}^3$ denotes the spatial position of the corresponding guidance point. For the 2D representation, we project $p_t$ onto a camera view $n\in\{1,\ldots,N_{\mathrm{cam}}\}$ to obtain the pixel-aligned Gaussian heatmap target $G^*_{t,n}$, directly identifying  interaction-relevant visual tokens but limited by camera FOV. For the 3D representation, the relative 3D offset target $\Delta p^*_t=p_t-p^{\mathrm{e}}_{t_0}$ is adopted, where $p^{\mathrm{e}}_{t_0}\in\mathbb{R}^3$ denotes the end effector position at the last observed frame $t_0$ before the target action chunk, describing the relative position of the end effector to the guidance point, providing complementary 3D geometric priors. Overall we jointly model 2D and 3D representations to guide action tokens toward interaction-relevant regions.

\noindent\textbf{Geometry-Guided Attention Modulation.} Since the relevant interaction target varies with the task instruction and execution state, GPVA integrates the proprioceptive state of robot and task instruction into visual tokens to jointly predict the 2D heatmap and corresponding 3D offset relative to the interaction target, which are further converted into attention-logit biases and added to unnormalized attention scores, guiding visual attention explicitly.


Spatiotemporal and semantic contexts are both required for predicting guidance points across future frames in the target chunk. At each GPVA layer, $N_{\mathrm{tok}}$ clean visual tokens from $N_{\mathrm{cam}}$ views at frame $t$ are first augmented with learned projections of robot state at $t_0$, temporal offset $t-t_0$, and normalized 2D coordinates, followed by LayerNorm to obtain $d_g$-dim state-aware features $v_t^{\mathrm{state}}\in\mathbb{R}^{N_{\mathrm{tok}}\times d_{g}}$. $v_t^{\mathrm{state}}$ is used to query instruction tokens $E_{text}$ through cross attention to emphasize interaction-relevant regions. An MLP is used to map updated features to {\footnotesize $v_t^{\mathrm{intr}}\in\mathbb{R}^{N_{\mathrm{tok}}\times d_{g}}$}.

The 2D branch decodes $v_t^{\mathrm{intr}}$ into the heatmap $\hat G_{t,n}$ indicating the location of the guidance point in each view.
We spatially aggregate the heatmap response score of each visual token producing $\rho_t^{\mathrm{2D}}\in(0,1)^{N_{\mathrm{tok}}}$, which is further converted into 2D attention-logit biases through $b_t^{\mathrm{2D}}=\log(\mathbf{1}+\rho_t^{\mathrm{2D}})$. Hence action queries are guided to concentrate on visual regions with stronger heatmap responses.
The 3D branch projects $v_t^{\mathrm{intr}}$ into the 3D geometric representation $\bar{v}_t^{\mathrm{intr}} \in \mathbb{R}^{d_{g}}$ along the dimension of token, jointly predicting the relative 3D offset $\Delta\hat p_t$ and 3D spatial uncertainty $\sigma_t$ that measures the confidence of the predicted offset. We project $(\Delta\hat p_t, \sigma_t)$ into a query that attends to $v_t^{\mathrm{intr}}$, yielding a spatial reference distribution $\rho_t^{\mathrm{3D}}\in\mathbb{R}^{N_{\mathrm{tok}}}$ over $N_{\mathrm{tok}}$ visual tokens. Taking the uniform weight $w_u=1/N_{\mathrm{tok}}$ as reference,  $\rho_t^{\mathrm{3D}}$ is converted into 3D attention-logit biases through $b_t^{\mathrm{3D}}=\tanh(\log(\rho_t^{\mathrm{3D}}/w_u))$ to amplify tokens above $w_u$ and suppress tokens below $w_u$, in which $\tanh$ is used to constrain $b_t^{\mathrm{3D}}$ to $(-1,1)^{N_{\mathrm{tok}}}$.

A learned gate $\beta_t\in(0,1)$, predicted from $\bar{v}_t^{\mathrm{intr}}$, adaptively combines  $b_t^{\mathrm{2D}}$ and $b_t^{\mathrm{3D}}$ into the unified bias $b_t$. 
$b_t$ is added to self attention logits before softmax to guide action queries toward guidance point associated tokens:
\begin{equation}
\label{eq:GPVA_injection}
b_t=\beta_t b_t^{\mathrm{2D}}+(1-\beta_t)b_t^{\mathrm{3D}}
\end{equation}

\noindent\textbf{Training and Inference.} Ground truth for guidance points is extracted from trajectories. \emph{More details are provided in the Supplementary Material.} For the 2D branch, $\mathcal{L}^{\mathrm{MSE}}_{t,n}=\|\hat G_{t,n}-G^*_{t,n}\|_2^2$ penalizes pixel-wise heatmap errors. Since MSE is insensitive to low-amplitude spurious peaks on distractors, we apply the Jensen-Shannon divergence~\cite{lin1991divergence} $\mathcal{L}^{\mathrm{JS}}_{t,n}$ between spatially normalized $\hat G_{t,n}$ and $G^*_{t,n}$ to align global distributions. Using the binary mask $m_{t,n}$ excluding targets outside the camera FOV, $\mathcal{L}_{\mathrm{2D}}$ is formulated as:
\begin{equation}
\label{eq:loss_2d_total}
\mathcal{L}_{\mathrm{2D}}=
\frac{\sum_{t,n}m_{t,n}\left(
\mathcal{L}^{\mathrm{MSE}}_{t,n}+
\lambda_{\mathrm{JS}}\mathcal{L}^{\mathrm{JS}}_{t,n}
\right)}{\sum_{t,n}m_{t,n}+\epsilon}
\end{equation}

For the 3D branch, the scale-normalized offset error is defined as $e_t=(\Delta\hat{p}_t-\Delta p^*_t)/\alpha_s$ where $\alpha_s$ denotes the offset scale, and  uncertainty objective is formulated as $e^\sigma_t = \sigma_t / \alpha_s - \operatorname{sg}(|e_t|)$. $\sigma_t$ is guided to estimate the magnitude of the offset error, and the stop-gradient operator $\operatorname{sg}(\cdot)$ prevents uncertainty supervision from perturbing offset regression learning. 2D and 3D branches provide complementary supervisions, which are trained jointly and combined into the overall GPVA objective $\mathcal{L}_{\mathrm{GPVA}}
=\lambda_{\mathrm{2D}}\mathcal{L}_{\mathrm{2D}}
+\lambda_{\mathrm{3D}}\mathcal{L}_{\mathrm{3D}}$:
\begin{equation}
\label{eq:loss_gpva}
\begin{aligned}
\mathcal{L}_{\mathrm{3D}}
&=\frac{1}{T}\sum_{t}\left(\|e_t\|_2^2 + \|e^\sigma_t\|_2^2\right)
\end{aligned}
\end{equation}
where $T$ denotes the number of predicted frames, $\lambda_{\mathrm{2D}}$ and $\lambda_{\mathrm{3D}}$ denote loss weights. $\mathcal{L}_{\mathrm{GPVA}}$ is added to video-action loss and optimized end-to-end with  transformer backbone. GPVA processes clean visual tokens under a structured causal mask during training, while operating on predicted visual tokens and updating the K/V cache with the context of real observations during inference.


\subsection{State-Conditioned Text Attention}
\label{sec:scta}



Accurate action generation requires relevant attendance to instruction semantics based on the current execution state. 
However, explicitly guiding instruction attention according to the evolving execution state remains widely overlooked.

\noindent\textbf{Execution State Formulation.} Let $h^\mu_{j,i}\in\mathbb{R}^D$ denote the hidden state of the $i$-th token of modality $\mu$ within the $j$-th chunk, where $\mu\in{\{v,a\}}$ indexes the video or action modality. The constraint $j<c$ is imposed, where $c$ denotes the target chunk, to prevent using current and future action targets in execution state aggregation during training. We project hidden states of both modalities into a shared $d_s$-dim space and aggregate tokens only from preceding chunks:
\begin{equation}
\begin{aligned}
y^{\mu}_{j,i}&=\operatorname{LN}^{\mu}\!\left(\mathcal{A}^\mu\!\left(h^{\mu}_{j,i}\right)\right)
\end{aligned}
\label{eq:scta-history-projection}
\end{equation}
where $\mathcal{A}^\mu$ denotes the modality-specific affine map from $\mathbb R^D$ to $\mathbb R^{d_s}$, and $\operatorname{LN}^{\mu}$ denotes modality-specific LayerNorm.

We collect the $y^\mu_{j,i}$ of historical video and action tokens from preceding chunks into $\mathcal{Y}_c$ for the target chunk $c$. These historical features are aggregated through multi-head attention using a learnable query $q_c^{\mathrm{state}}\in\mathbb{R}^{d_s}$ shared across chunks. A state head then combines the attention output with $q_c^{\mathrm{state}}$ to produce the execution state $z_c$:
\begin{equation}
z_c=\Phi\!\left(\left[q_{c}^{\mathrm{state}};\operatorname{MHA}\!\left(q_{c}^{\mathrm{state}},\mathcal Y_c,\mathcal Y_c\right)\right]\right)
\label{eq:scta-execution-state}
\end{equation}
The state head $\Phi:\mathbb{R}^{2d_s}\to\mathbb{R}^{d_s}$ maps the concatenation of $q_c^{\mathrm{state}}$ and the attention output to the execution state $z_c$. A learnable relative chunk-distance bias is further incorporated into attention logits, enabling the capability of distinguishing recent from distant historical context.

\noindent\textbf{Instruction-Guided Action Modulation.} $N_{text}$ instruction token embeddings in $E_{text}$ are averaged to obtain the global task summary, which is integrated with the state $z_c$ to form a query that attends to $E_{text}$ through multi-head attention retrieving state-conditioned text semantics. State-conditioned text semantics and the global task summary jointly determine  the ``scale and shift'' parameters applied on noisy action features through FiLM.

Let $h^a_{\ell,c,i}$ denote the hidden state of the $i$-th noisy action token in the chunk $c$ at the layer $\ell$. FiLM modulates $h^a_{\ell,c,i}$ through the following residual update:
\begin{equation}
h^{a\prime}_{\ell,c,i}
=
h^a_{\ell,c,i}
+s_{\ell,c}\odot\operatorname{LN}(h^a_{\ell,c,i})
+b_{\ell,c}
\label{eq:scta-film}
\end{equation}
where $\odot$ denotes element-wise multiplication. The scale $s_{\ell,c}$ and shift $b_{\ell,c}$ are shared across action tokens in chunk $c$, and FiLM is applied only on the noisy action stream.

\noindent\textbf{Interval-Constrained Entropy Loss.}\ 
We introduce an auxiliary entropy loss to encourage appropriate attention modulation over instruction embeddings according to the execution stage. The auxiliary loss is designed to constrain the normalized entropy of the action-to-text attention distribution $\boldsymbol{\pi}$, discouraging both dispersive and overly concentrated attention distribution:
\begin{equation}
\begin{gathered}
\bar{\mathcal{H}}(\boldsymbol{\pi})
= -\frac{1}{\log N_{text}}
  \sum_{i=1}^{N_{text}}\pi_i\log\pi_i \\[2pt]
\scalebox{0.86}{$\displaystyle
\mathcal{L}_{\mathrm{ent}}
= \mathbb{E}\Bigl[
  \operatorname{ReLU}\!\bigl(
    \bar{\mathcal{H}}(\boldsymbol{\pi})-\tau_{\mathrm{high}}
  \bigr)^2+
  \operatorname{ReLU}\!\bigl(
    \tau_{\mathrm{low}}-\bar{\mathcal{H}}(\boldsymbol{\pi})
  \bigr)^2
  \Bigr]
$}
\end{gathered}
\label{eq:scta-entropy}
\end{equation}
where $\pi_i$ denotes the softmax-normalized attention weight assigned to the $i$-th instruction token, and $N_{text}$ denotes the number of instruction tokens.  $\tau_{\mathrm{low}}$ and $\tau_{\mathrm{high}}$ specify lower and upper bounds on the normalized entropy. The penalty is zero when the normalized entropy falls within the interval.

\begin{table*}[t]
  \centering
  \caption{Performance on the LIBERO-X benchmark. The success rate (\%) is adopted for evaluation.}
  \vspace{-0.3cm}
  \label{tab:liberox_main}
  \small
  \setlength{\tabcolsep}{4pt}
  \begin{tabular*}{\textwidth}{@{\extracolsep{\fill}}l*{8}{c}@{}}
    \toprule
    Level & OpenVLA-OFT~\cite{kim2025openvlaoft} & X-VLA~\cite{zhengxvla2025} & $\pi_{0}$~\cite{black2024pi0}  & GR00T1.5~\cite{bjorck2025gr00t} &  $\pi_{0.5}$~\cite{black2025pi05} & LingBot-VA~\cite{li2026lingbotva} & Baseline & \textbf{AG-WAM} \\
    \midrule
    L1 & 29.0\% & 30.1\% & 29.4\%  & 43.3\%  & 65.2\% & 68.8\% & 71.6\% & \bf{79.2\%} \\
    L2 & 17.6\% & 22.6\% & 21.9\% & 32.9\%  & 53.2\% & 59.0\% & 60.3\% & \bf{69.0\%} \\
    L3 & 8.8\% & 10.3\% & 11.0\%  & 18.7\%  & 36.0\% & 36.2\% & 38.7\% & \bf{44.3\%} \\
    L4 & 6.4\% & 6.0\% & 7.6\% & 13.3\%  & 24.1\% & 27.8\% & 29.1\% & \bf{33.5\%} \\
    L5 & 4.2\% & 4.1\% & 5.1\% & 9.7\%  & 18.0\% & 19.7\% & 20.2\% & \bf{24.8\%} \\
    \textbf{Average} & 13.2\% & 14.6\% & 15.0\% & 23.6\%  & 39.3\% & 42.3\% & 44.0\% & \bf{50.2\%} \\
    \bottomrule
  \end{tabular*}
\end{table*}

\begin{table*}[t]
  \centering
  \caption{Performance on the LIBERO-Plus benchmark. The success rate (\%) is adopted for evaluation.}
  \vspace{-0.3cm}
  \label{tab:liberoplus_main}
  \small
  \setlength{\tabcolsep}{4pt}
  \begin{tabular*}{\textwidth}{@{\extracolsep{\fill}}l*{9}{c}@{}}
    \toprule
    Method & Plus-SFT & Camera & Robot & Language & Light & Background & Noise & Layout & \textbf{Overall} \\
    \midrule
   OpenVLA-OFT~\cite{kim2025openvlaoft} & $\times$ & 56.4\% & 31.9\% & 79.5\% & 88.7\% & 93.3\% & 75.8\% & 74.2\% & 69.6\% \\
   StarVLA-OFT~\cite{starvla2026} & $\times$ & 47.0\% & 60.1\% & 87.0\% & 96.3\% & 95.3\% & 73.1\% & 79.2\% & 75.0\% \\
   Cosmos-Policy~\cite{kim2026cosmospolicy} & $\times$ & 75.8\% & 63.3\% & 81.7\% & 96.5\% & 88.9\% & 92.7\% & 82.2\% & 82.2\% \\
   Being-H0.7~\cite{luobeingh2026} & $\times$ & 82.0\% & 59.0\% & 82.8\% & 97.8\% & 90.0\% & 93.5\% & 88.5\% & 84.1\% \\
   InternVLA-A1.5~\cite{ma2026internvla} & $\times$ & 83.1\% & 55.1\% & 86.9\% & 96.4\% & 98.2\% & 95.6\% & 85.2\% & 84.8\% \\
   OA-WAM~\cite{liuoawam2026} & $\times$ & 80.5\% & 89.6\% & 85.3\% & 96.5\% & 95.9\% & 75.6\% & 82.8\% & 85.7\% \\
   $\pi_{0.5}$~\cite{black2025pi05} & $\times$ & 75.8\% & 79.4\% & 83.3\% & 95.5\% & 95.0\% & 89.6\% & 87.0\% & 85.7\% \\
   Kairos-joint~\cite{team2026kairos} & \checkmark & 95.9\% & 74.6\% & \bf{95.3\%} & 97.1\% & 97.1\% & 95.4\% & 83.8\% & 90.8\% \\
   \scriptsize{Qwen-RobotManip-Context~\cite{qwenrobotmanip2026}} & $\times$ & 89.9\% & 83.9\% & 86.5\% & 98.6\% & \bf{99.9\%} & 97.9\% & 87.5\% & 91.4\% \\
   ZimaBlue~\cite{wu2026zimablue} & \checkmark & 95.4\% & 81.1\% & 88.8\% & \bf{98.8\%} & 99.2\% & 96.5\% & 84.3\% & 91.5\% \\
   \textbf{AG-WAM\_zeroshot} & $\times$ & 93.2\% & 89.7\% & 88.1\% & 96.8\% & 98.2\% & 95.3\% & 84.2\% & 91.8\% \\
   \textbf{AG-WAM} & \checkmark & \bf{98.4\%} & \bf{94.5\%} & 90.1\% & 98.1\% & 99.0\% & \bf{98.4\%} & \bf{88.9\%} & \bf{95.1\%} \\
    \bottomrule
  \end{tabular*}
\end{table*}


\begin{table*}[!t]
  \centering
  \caption{Performance on the RoboTwin-2.0 benchmark. The success rate (\%) is adopted for evaluation.}
  \vspace{-0.3cm}
  \label{tab:robotwin_main}
  \footnotesize
  \setlength{\tabcolsep}{4pt}
  \begin{tabular*}{\textwidth}{@{\extracolsep{\fill}}l*{10}{c}@{}}
    \toprule
     & $\pi_{0.5}$ & Motus & Being-H0.7 & Fast-WAM & InternVLA-A1.5 & LingBotVA-2.0 & \tiny{Qwen-RobotManip-Context}  & MotuBrain & Kairos & \textbf{AG-WAM} \\
    Setting & \cite{black2025pi05} & \cite{bi2026motus}  & \cite{luobeingh2026} & \cite{yuan2026fastwam} & \cite{ma2026internvla} & \cite{zhang2026lingbotva2} & \cite{qwenrobotmanip2026} & \cite{xiang2026motubrain} & \cite{team2026kairos} &  \\
    \midrule
    Clean & 82.74\% & 88.66\%  & 90.20\% & 91.88\% & 93.30\% & 93.80\% & 93.70\% & 95.80\% & 96.90\% & \bf{97.06\%} \\
    Random & 76.76\% & 87.02\%  & 89.60\% & 91.78\% & 93.00\% & 93.40\% & 94.00\% & 96.08\% & 95.20\% & \bf{97.22\%} \\
    \textbf{Average} & 79.75\% & 87.84\%  & 89.90\% & 91.83\% & 93.15\% & 93.60\% & 93.85\% & 95.94\% & 96.05\% & \bf{97.14\%} \\
    \bottomrule
  \end{tabular*}
  
  \par\vspace{1.5em}
  \noindent\begin{minipage}[t]{0.48\textwidth}
  \centering
  \captionsetup{font={footnotesize}}
  \caption{Ablation studies on LIBERO-X. Success rate (\%) is adopted.}
  \vspace{-0.3cm}
  \label{tab:liberox_ablation}
  \small
  \setlength{\tabcolsep}{4pt}
  \begin{tabular*}{\linewidth}{@{\extracolsep{\fill}}l*{5}{c}@{}}
    \toprule
    Method & L1 & L2 & L3 & L4 & L5 \\
    \midrule
    Baseline(BS) & 71.6\% & 60.3\% & 38.7\% & 29.1\% & 20.2\% \\
    BS + CAPA & 72.1\% & 60.5\% & 39.0\% & 29.2\% & 20.5\% \\
    BS + SCTA & 74.7\% & 63.8\% & 41.2\% & 30.4\% & 22.1\% \\
    BS + GPVA & 76.9\% & 65.9\% & 42.8\% & 32.7\% & 23.7\% \\
    \scriptsize{BS + SCTA + GPVA} & 78.8\% & 68.5\% & 43.8\% & 33.2\% & 24.5\% \\
    \textbf{AG-WAM} & \bf{79.2\%} & \bf{69.0\%} & \bf{44.3\%} & \bf{33.5\%} & \bf{24.8\%} \\
    \bottomrule
  \end{tabular*}

  \end{minipage}\hfill
  \begin{minipage}[t]{0.48\textwidth}
  \centering
  \caption{Efficiency analysis on LIBERO-X. We implement AG-WAM using PyTorch without any acceleration operations. The inference latency is measured per action chunk.}
  \vspace{-0.3cm}
  \label{tab:liberox_efficiency}
  \small
  \setlength{\tabcolsep}{2pt}
  \begin{tabular*}{\linewidth}{@{\extracolsep{\fill}}l*{3}{c}@{}}
    \toprule
    Method & \shortstack{Overall\\Success\\Rate} & \shortstack{Training\\Efficiency\\(s/iter)} &  \shortstack{Inference\\Latency\\(ms)} \\
    \midrule
    Baseline(BS) & 44.0\% & 16.7($1.00\times$) & 1585.6($1.00\times$) \\
    AG-WAM w/o CAPA & 49.8\% & 18.4($0.91\times$) & 1722.5($0.92\times$) \\
    w/ CAPA $\eta_{\mathrm{ret}}=0.7$ & 50.1\% & 12.9($1.30\times$) & 1230.4($1.29\times$) \\
    w/ CAPA $\eta_{\mathrm{ret}}=\mathbf{0.6}$ & \underline{50.2\%} & \underline{10.1($1.65\times$)} & \underline{1090.2($1.45\times$)} \\
    w/ CAPA $\eta_{\mathrm{ret}}=0.5$ & 48.1\% & 8.6($1.94\times$) & 1019.2($1.56\times$) \\
    \bottomrule
  \end{tabular*}
  \end{minipage}
\end{table*}

\begin{figure*}[t]
\centering
\includegraphics[width=\textwidth]{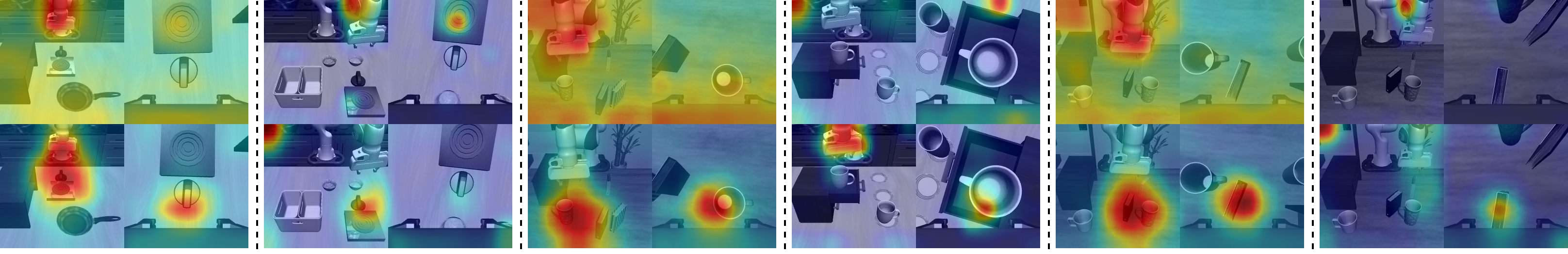}
\vspace{-0.7cm}
\captionsetup{font={footnotesize}}
\caption{Visualized effectiveness of GPVA. Top: dispersive or misplaced visual attention without GPVA. Bottom: guided visual attention with GPVA.}
\label{fig:Fig3}
\end{figure*}
\section{Experiments}
\label{sec:experiments}

We evaluate our approach on three public robotic manipulation benchmarks: LIBERO-X~\cite{wang2026liberox}, LIBERO-Plus~\cite{fei2025liberoplus}, and RoboTwin-2.0~\cite{chen2025robotwin2}. Our AG-WAM establishes new state-of-the-art performance on all three benchmarks, meanwhile significantly improving training and inference efficiency. \emph{More experiments are provided in  Supplementary Material.}


\subsection{Datasets}
\label{sec:exp_benchmarks}

\noindent\textbf{LIBERO-X.}
LIBERO-X extends saturated LIBERO~\cite{liu2023libero} with five evaluation levels covering various spatial, object, and linguistic shifts. Performance of existing policies degrades severely on this latest challenging benchmark. 




\smallskip\noindent\textbf{LIBERO-Plus.}
LIBERO-Plus contains over $10{,}000$ evaluation variants across seven perturbation dimensions: layout, camera, state, language, lighting, background, and noise. 


\smallskip\noindent\textbf{RoboTwin-2.0.}
RoboTwin-2.0 evaluates $50$ bimanual manipulation tasks under clean and randomized settings, randomized over clutter, visual appearance, and instructions. 

\smallskip\noindent\textbf{Evaluation Protocol.}
Following~\cite{li2026lingbotva}, we report the success rate (\textbf{SR}, \%) of robot manipulation. 


\subsection{Implementation Details}
\label{sec:exp_implementation}



CAPA is inserted before the transformer backbone. $\eta_{\mathrm{ret}}$ is set to $0.6$ indicating that $40\%$ of video tokens are discarded. $10\%$ of all video tokens are randomly sampled for retention. The transformer backbone ($30$ layers in total, $D$ is $3072$) processes the retained video tokens together with all action tokens. 
Both GPVA and SCTA use retained video tokens by CAPA, and GPVA aligns heatmap responses with these retained video tokens based on their original spatial positions.
GPVA and SCTA are inserted in the transformer backbone at layers $\{1, 8, 15, 22, 29\}$ with shared parameters across layers. $d_{g}$ in GPVA is set to $256$, $d_s$ in SCTA is set to $512$, and $d_r$ in CAPA is set to $512$.


\smallskip\noindent\textbf{Training and Inference.}
We train the entire model end-to-end in bfloat16 with FSDP using AdamW ($\beta_1 = 0.9, \beta_2 = 0.95$), learning rate $1 \times 10^{-5}$, and CFG dropout $0.1$. In  $\mathcal{L}_{\mathrm{GPVA}}$, $\lambda_{\mathrm{JS}}$ is set to $0.1$, $\lambda_{\mathrm{2D}}$ and $\lambda_{\mathrm{3D}}$ are both set to $1.0$. In $\mathcal{L}_{\mathrm{ent}}$, $\tau_{\mathrm{low}}$ is set to $0.2$, and $\tau_{\mathrm{high}}$ is set to $0.6$. The training is conducted on $16$ NVIDIA-80GB-A800 GPUs with a total batch size of $32$ for $100$K steps on each dataset. For clarification, we only use publicly available training data, without using any extra private embodied data. In RoboTwin-2.0, both clean and randomized training data is used. The inference is conducted on a single NVIDIA-24GB-4090 GPU. 


\subsection{Evaluations on Manipulation Benchmarks}
\label{sec:exp_main_results}

\begin{figure*}[!t]
\centering
\includegraphics[width=\textwidth]{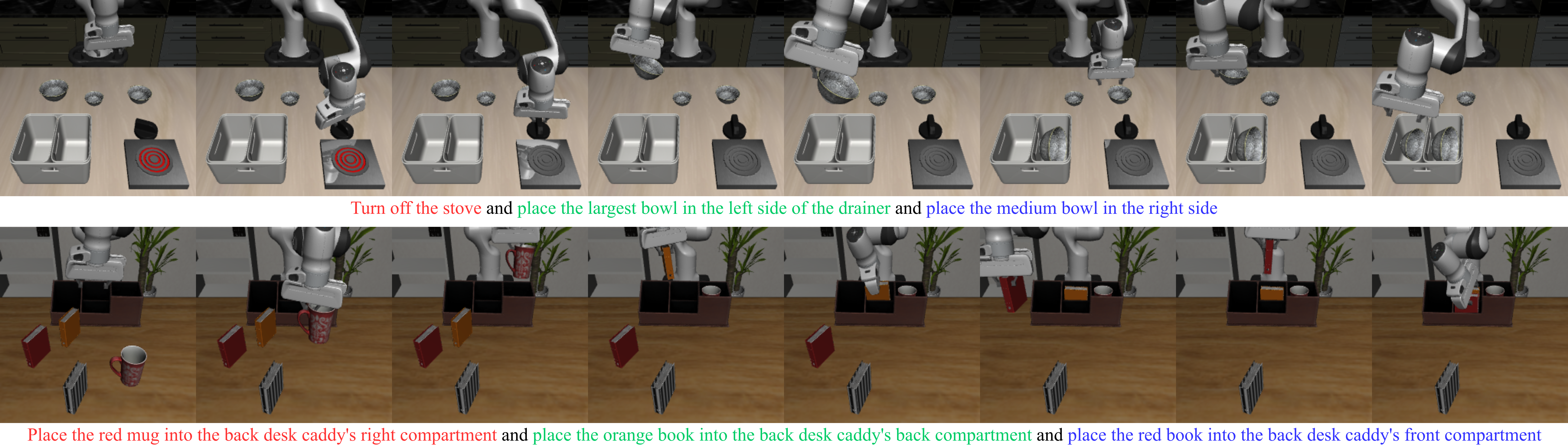}
\vspace{-0.7cm}
\caption{Qualitative results of AG-WAM on challenging LIBERO-X. Complex long-horizon manipulation tasks are successfully handled.}
\label{fig:Fig5}
\end{figure*}

As shown in Tab.~\ref{tab:liberox_main}, on the LIBERO-X benchmark, AG-WAM significantly outperforms all other methods by a notable margin, especially the previous best LingBot-VA~\cite{li2026lingbotva} by $10.4\%$, $10.0\%$, $8.1\%$, $5.7\%$, and $5.1\%$ at levels of L1$\sim$L5 respectively. We further provide qualitative results of our method on complex long-horizon manipulation tasks, as shown in Fig.~\ref{fig:Fig5}. The strong capabilities of our end-to-end framework with explicit and unified attention guidance are demonstrated on this latest challenging benchmark. On the LIBERO-Plus benchmark, as shown in Tab.~\ref{tab:liberoplus_main}, our method achieves a remarkable performance of $95.1\%$ overall SR. In particular, AG-WAM outperforms the most competitive methods ZimaBlue~\cite{wu2026zimablue} and Kairos-joint~\cite{team2026kairos} by $3.6\%$ and $4.3\%$ respectively, which reveals remarkable robustness performance of our method. On the RoboTwin-2.0 benchmark, as shown in Tab.~\ref{tab:robotwin_main}, our method achieves the highest average success rate of $97.14\%$. Particularly, without using any private embodied training data, AG-WAM outperforms the best competitors Kairos~\cite{team2026kairos} and  MotuBrain~\cite{xiang2026motubrain} that use massive private embodied training data by $1.1\%$ and $1.2\%$ respectively. Our AG-WAM is the first method achieving success rates of over $97\%$ on both settings. 



\begin{figure}[!t]
\centering
\includegraphics[width=0.5\textwidth]{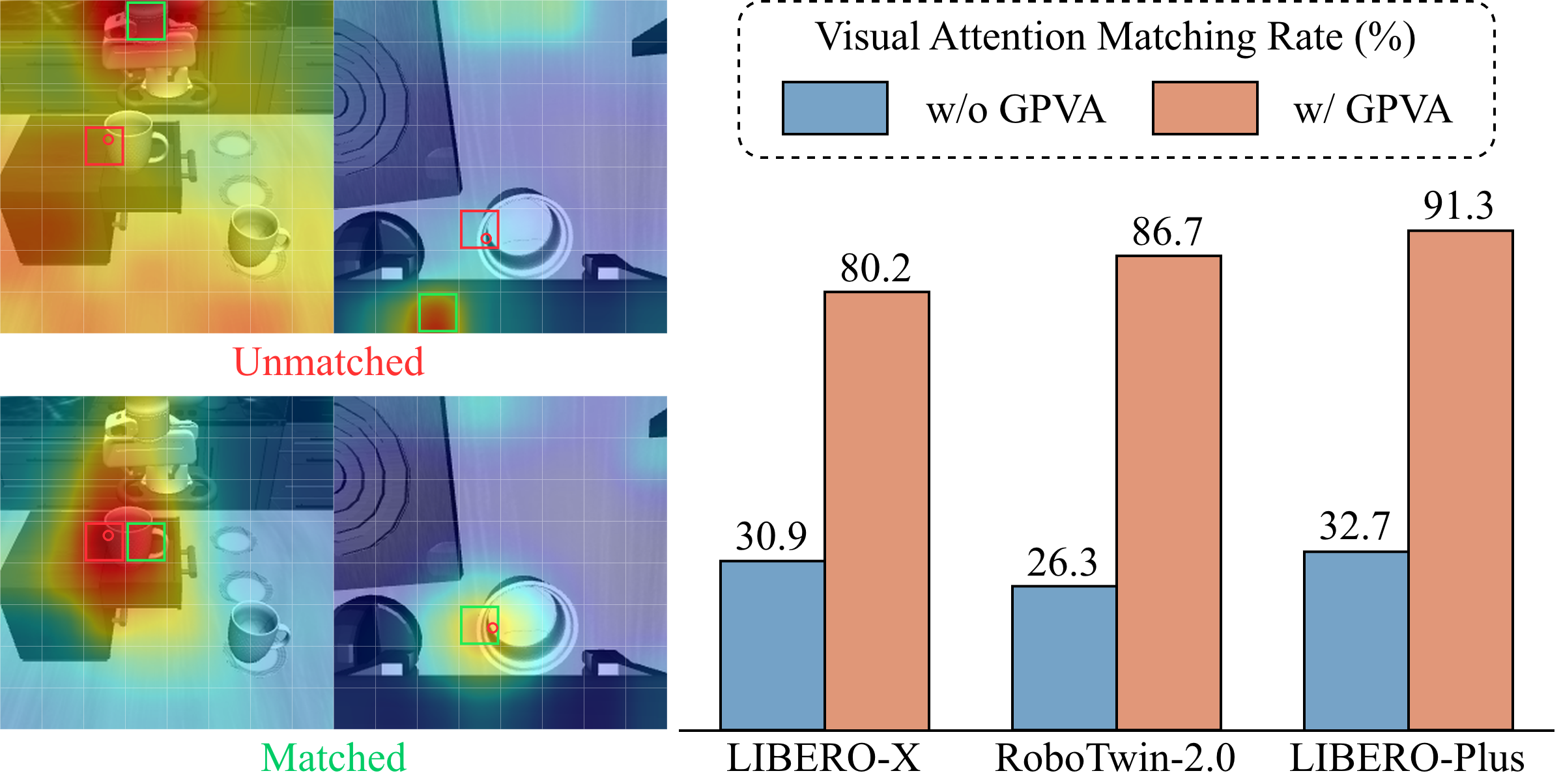}
\vspace{-0.7cm}
\caption{The effectiveness of GPVA on the matching rate (\%) of visual attention. The matched and unmatched examples are displayed in the left column, where red and green boxes represent the ground truth and prediction respectively. }
\label{fig:Fig6}
\end{figure}

\begin{figure}[!t]
\centering
\includegraphics[width=0.5\textwidth]{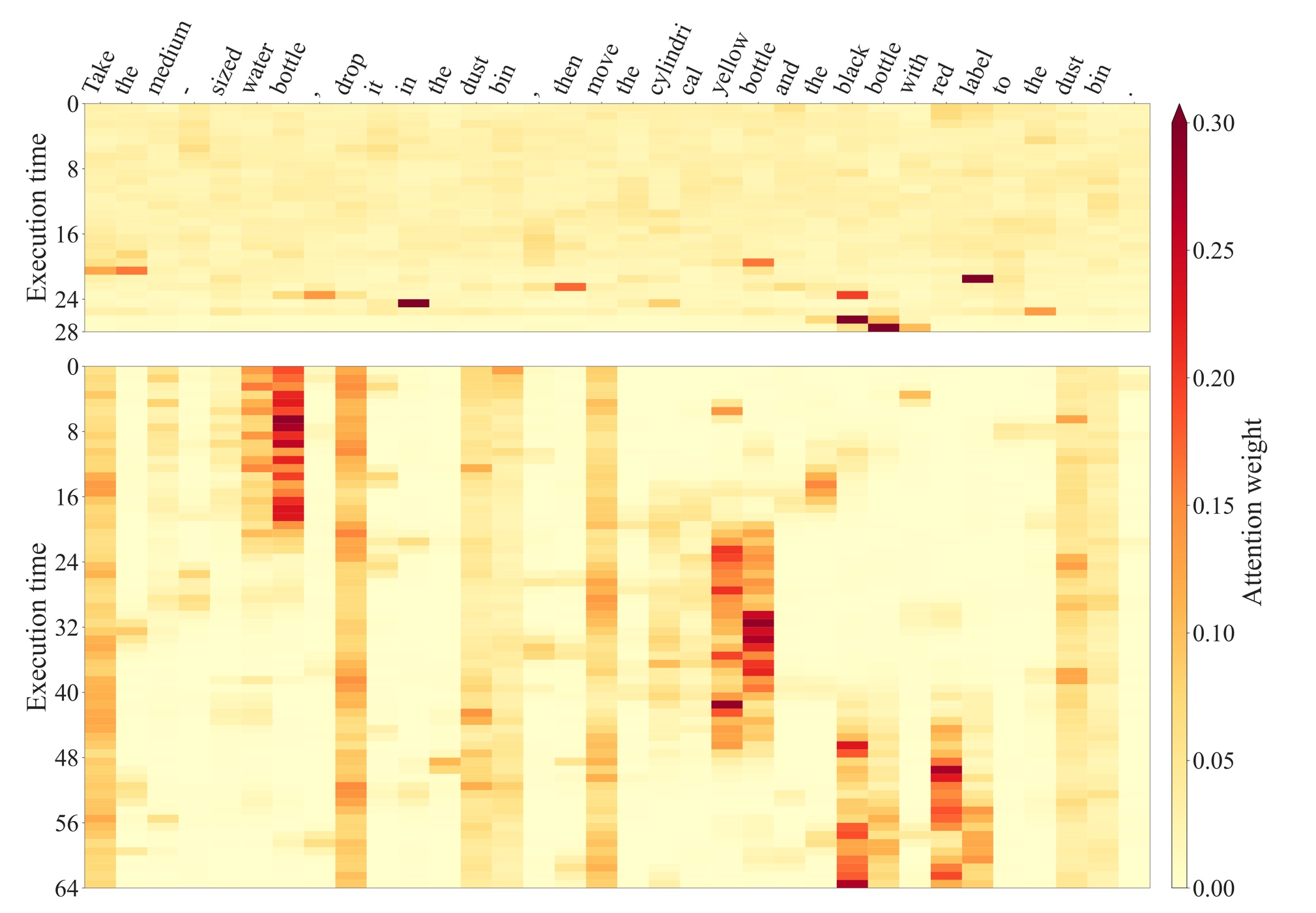}
\vspace{-0.7cm}
\caption{Visualized effectiveness of SCTA. Top: dispersive attention with erroneous activations without SCTA. Bottom: guided state-dependent instruction attention with SCTA.}
\label{fig:Fig4}
\end{figure}

\subsection{Ablation Studies}
\label{sec:ablation}

We conduct detailed ablation studies on the challenging LIBERO-X dataset. 
As shown in Tab.~\ref{tab:liberox_ablation}, our baseline model achieves highly competitive performance. Applying CAPA with $40\%$ of video tokens discarded maintains overall performance stable, meanwhile significantly improving training and inference efficiency as shown in Tab.~\ref{tab:liberox_efficiency}.
In addition, by integrating SCTA, success rates are effectively improved by $3.1\%$, $3.5\%$, $2.5\%$, $1.3\%$, and $1.9\%$ at levels of L1$\sim$L5 respectively.
As shown in Fig.~\ref{fig:Fig4}, we also provide visualized effectiveness evidence of SCTA, where the top row indicates dispersively distributed instruction attention and activations on tokens erroneously associated with later goals, and the bottom row indicates instruction attention shifting toward relevant segments as execution progresses. Furthermore, integrating GPVA can notably improve success rates by $5.3\%$, $5.6\%$, $4.1\%$, $3.6\%$, and $3.5\%$ at levels of L1$\sim$L5 respectively.
As shown in Fig.~\ref{fig:Fig3}, we additionally provide visualized effectiveness evidence of GPVA, where the top row indicates dispersive or misplaced visual attention on irrelevant or erroneous regions without GPVA, and the bottom row indicates guided concentrating visual attention on interaction-relevant regions with GPVA. As shown in Fig.~\ref{fig:Fig6}, we further demonstrate the effectiveness of GPVA by evaluating the matching rate of visual attention, where the average matching rate on three datasets is improved from $30.0\%$ to $86.1\%$. Overall, integrating three attention guidance modules, AG-WAM achieves state-of-the-art performance on LIBERO-X, with improvements of $7.6\%$, $8.7\%$, $5.6\%$, $4.4\%$, and $4.6\%$ at levels of L1$\sim$L5 over the baseline.

\subsection{Training and Inference Efficiency Analysis}
\label{sec:Training_Inference}

We conduct training and inference efficiency analysis on the LIBERO-X dataset.
As shown in Tab.~\ref{tab:liberox_efficiency}, applying CAPA with $40\%$ of video tokens discarded delivers significant improvements in both training and inference efficiency, achieving a $1.65\times$ training speedup, a $1.45\times$ inference speedup, and a relative $20.8\%$ training memory reduction from $62.9$ GB to $49.8$ GB. Experimental results demonstrate the effectiveness of predictive video token selection and corresponding informative attention guidance. 
\emph{In this work, we do not compare absolute latency with methods specifically designed for lightweight architectures.} CAPA can be easily integrated into other lightweight WAMs, offering considerable potential for further acceleration.


\section{Conclusion}
\label{sec:conclusion}

In this paper, we address the problem of implicit attention allocation in existing WAMs. Accordingly, we propose an end-to-end framework termed AG-WAM integrating three explicit attention guidance modules. Experimental results demonstrate the superiority of our method over other state-of-the-art methods on multiple robotic manipulation benchmarks. In the future, we will explore adapting AG-WAM to other architectures, such as lightweight WAMs.

\section*{LLM Usage Statement}
\label{sec:llm}

No LLMs or generative AI tools were used in this work.
{
    \small
    \bibliographystyle{ieeenat_fullname}
    \bibliography{main}
}

\clearpage
\setcounter{page}{1}
\maketitlesupplementary

\begin{algorithm*}[t]
\caption{Single-Arm Offline Guidance Point Extraction (Applied Per Arm in Bimanual Setups)}
\label{alg:guidance_extraction}
\footnotesize
\begin{algorithmic}[1]
\REQUIRE Trajectory $\{(p_t^{\mathrm{e}}, o_t)\}_{t=1}^{T_{\mathrm{ep}}}$ of length $T_{\mathrm{ep}}$  (end-effector 3D position $p_t^{\mathrm{e}} \in \mathbb{R}^3$ and gripper opening  state $o_t$); gripper thresholds $\tau_{\mathrm{close}} < \tau_{\mathrm{open}}$; minimum duration $L_{\min}$; velocity drop ratio $\gamma_{\mathrm{vel}} \in (0,1)$ and minimum speed $v_{\min}^{\mathrm{e}}$; temporal margin $\Delta_{\min}$; minimum displacement threshold $d_{\min}$
\ENSURE Per-frame 3D guidance points $\{p_t\}_{t=1}^{T_{\mathrm{ep}}}$

\STATE \textbf{// Step 1: Extract gripper-based grasp-release events}
\STATE Detect $M \ge 0$ grasp-release pairs $\{(t_{\mathrm{grasp}}^{(m)}, t_{\mathrm{release}}^{(m)})\}_{m=1}^{M}$ where $o_t < \tau_{\mathrm{close}}$ for $\ge L_{\min}$ frames, bounded by $o_t > \tau_{\mathrm{open}}$
\STATE $\mathcal{T} \leftarrow \bigcup_{m=1}^{M} \{t_{\mathrm{grasp}}^{(m)}, t_{\mathrm{release}}^{(m)}\}$ \COMMENT{Initialized event-timestamp set ($\mathcal{T} = \emptyset$ if $M=0$)}

\STATE \textbf{// Step 2: Supplement with kinematic change-points outside carry intervals}
\STATE Compute smoothed speed $\{v_t^{\mathrm{e}}\}_{t=1}^{T_{\mathrm{ep}}}$ via frame-to-frame finite differences of $\{p_t^{\mathrm{e}}\}_{t=1}^{T_{\mathrm{ep}}}$ followed by moving-average smoothing
\STATE Find the first maximal interval $\mathcal{I}_{\mathrm{init}} = [t_{\mathrm{onset}}, t_{\mathrm{stable}}]$ of length $\ge L_{\min}$ such that $v_t^{\mathrm{e}} > v_{\min}^{\mathrm{e}}$ for all $t \in \mathcal{I}_{\mathrm{init}}$
\IF{$\mathcal{I}_{\mathrm{init}}$ exists}
    \STATE Estimate baseline speed $v_{\mathrm{init}}^{\mathrm{e}}$ as the median of $v_t^{\mathrm{e}}$ over $\mathcal{I}_{\mathrm{init}}$
    \STATE Partition frames outside carry intervals $\bigcup_{m=1}^{M} [t_{\mathrm{grasp}}^{(m)}, t_{\mathrm{release}}^{(m)}]$ into contiguous non-carry segments
    \FOR{each non-carry segment $\mathcal{J}$}
        \STATE $\mathcal{C} \leftarrow \emptyset$ \COMMENT{Candidate timestamps for $\mathcal{J}$}
        \STATE $\mathcal{I}_{\mathrm{slow}} \leftarrow$ maximal sub-intervals $[t_{\mathrm{start}}^{(j)}, t_{\mathrm{end}}^{(j)}] \subseteq \mathcal{J}$ ($t_{\mathrm{start}}^{(j)} > t_{\mathrm{onset}}$, length $\ge L_{\min}$) with $v_t^{\mathrm{e}} < \gamma_{\mathrm{vel}}\,v_{\mathrm{init}}^{\mathrm{e}}$
        \FOR{each $[t_{\mathrm{start}}^{(j)}, t_{\mathrm{end}}^{(j)}] \in \mathcal{I}_{\mathrm{slow}}$}
            \IF{$\|p_{t_{\mathrm{end}}^{(j)}}^{\mathrm{e}} - p_{t_{\mathrm{start}}^{(j)}}^{\mathrm{e}}\|_2 > d_{\min}$}
                \STATE $\mathcal{C} \leftarrow \mathcal{C} \cup \{t_{\mathrm{start}}^{(j)}, t_{\mathrm{end}}^{(j)}\}$
            \ELSE
                \STATE $\mathcal{C} \leftarrow \mathcal{C} \cup \{\lfloor (t_{\mathrm{start}}^{(j)} + t_{\mathrm{end}}^{(j)}) / 2 \rfloor\}$
            \ENDIF
        \ENDFOR
        \STATE $\mathcal{M} \leftarrow \bigl\{c \in \mathcal{C} \;\big|\; \mathcal{T} = \emptyset \text{ or } \min_{t\in\mathcal{T}} |c - t| > \Delta_{\min}\bigr\}$
        \IF{$\mathcal{M} \neq \emptyset$}
            \IF{$\mathcal{J}$ ends at $T_{\mathrm{ep}}$}
                \STATE $\mathcal{T} \leftarrow \mathcal{T} \cup \{\min(\mathcal{M})\}$ \COMMENT{Terminal segment: keep onset only; endpoint handled in Step~3}
            \ELSE
                \STATE $t_{\mathrm{first}} \leftarrow \min(\mathcal{M}), \quad t_{\mathrm{last}} \leftarrow \max(\mathcal{M})$ \COMMENT{Non-terminal segment: onset \& completion}
                \STATE $\mathcal{T} \leftarrow \mathcal{T} \cup \{t_{\mathrm{first}}\}$
                \IF{$t_{\mathrm{last}} - t_{\mathrm{first}} > \Delta_{\min}$}
                    \STATE $\mathcal{T} \leftarrow \mathcal{T} \cup \{t_{\mathrm{last}}\}$ \COMMENT{Keep completion for extended interaction}
                \ENDIF
            \ENDIF
        \ENDIF
    \ENDFOR
\ENDIF

\STATE \textbf{// Step 3: Terminal fallback \& stage-wise assignment}
\IF{$\mathcal{T} = \emptyset$}
    \STATE $\mathcal{T} \leftarrow \{T_{\mathrm{ep}}\}$
\ELSIF{$T_{\mathrm{ep}} - \max(\mathcal{T}) > \Delta_{\min}$ \AND $\|p_{T_{\mathrm{ep}}}^{\mathrm{e}} - p_{\max(\mathcal{T})}^{\mathrm{e}}\|_2 > d_{\min}$}
    \STATE $\mathcal{T} \leftarrow \mathcal{T} \cup \{T_{\mathrm{ep}}\}$ \COMMENT{Capture terminal push/twist until success}
\ENDIF
\STATE Sort $\mathcal{T} = \{t_1 < \dots < t_K\}$ ($K = |\mathcal{T}| \ge 1$)
\FOR{$t = 1$ \TO $T_{\mathrm{ep}}$}
    \STATE Find the earliest event $t_k \in \mathcal{T}$ such that $t \le t_k$ (or $t_K$ if $t > t_K$)
    \STATE $p_t \leftarrow p_{t_k}^{\mathrm{e}}$
\ENDFOR
\RETURN $\{p_t\}_{t=1}^{T_{\mathrm{ep}}}$
\end{algorithmic}
\end{algorithm*}

\section{Additional Details and Explanations}

\subsection{The Ground-Truth Extraction of Guidance Point in GPVA}
\label{sec:supp_gpva_extraction}

As described in Sec.~\ref{sec:GPVA} in the main paper, GPVA supervises each future frame $t$ in the target action chunk using the guidance point $p_t \in \mathbb{R}^3$, from which both the pixel-aligned 2D Gaussian heatmap target $G_{t,n}^*$ and the relative 3D offset target $\Delta p_t^* = p_t - p_{t_0}^{\mathrm{e}}$ are derived. \Cref{alg:guidance_extraction} details our offline pipeline for extracting per-frame guidance points $\{p_t\}_{t=1}^{T_{\mathrm{ep}}}$ from a trajectory of length $T_{\mathrm{ep}}$.

\paragraph{Implementations.}
Alg.~\ref{alg:guidance_extraction} illustrates our offline extraction pipeline for single-arm manipulation. For bimanual setups, the same procedure is applied independently to each arm. Specifically, for bimanual supervisions in GPVA, the extracted left-arm and right-arm guidance points are processed by separate  prediction heads in both 2D and 3D branches: the 2D branch predicts the per-arm Gaussian heatmap across all three camera views arranged in the T-shaped composite input, and the 3D branch predicts the per-arm relative 3D offset and spatial uncertainty. The resulting attention biases jointly guide action queries toward interaction targets of both arms. 

In Alg. ~\ref{alg:guidance_extraction}, Step $1$ first identifies grasp-release transitions (\emph{e.g.} picking up a bowl and placing it onto a plate) directly from gripper opening state changes. Step $2$ then supplements these events with kinematic deceleration points outside carry intervals to capture non-prehensile interactions (\emph{e.g.} pressing a button, twisting a knob, or pulling open a cabinet door or drawer before grasping). Step $3$ finally handles terminal interactions (\emph{e.g.} closing a microwave door until task completion) and assigns each frame $t$ the 3D position of its upcoming key event.



To reduce the sensitivity to jitters in proprioceptive and gripper signals, Alg.~\ref{alg:guidance_extraction} uses separate gripper closing and opening thresholds ($\tau_{\mathrm{close}} < \tau_{\mathrm{open}}$) to identify the states of grasp and release, and smooths the end-effector speed with a moving average to reduce false event detections, meanwhile the displacement threshold ($d_{\min}$) is adopted to select event candidates and the minimum temporal separation ($\Delta_{\min}$) limits redundant event detections. For all models with GPVA, we randomly perturb camera calibrations in $20\%$ of training samples during training. 
Combined with the designed soft 2D Gaussian heatmap and uncertainty-aware 3D offset regression, GPVA is inherently insensitive to camera calibration errors.

During training, feeding clean visual tokens to GPVA follows the standard teacher-forcing paradigm, governed by the structured causal mask to prevent the information leakage. At inference, clean visual tokens are replaced with predicted visual tokens predicting future visual observations prior to the action generation, which guarantees no information leakage, while requiring neither the camera calibration as input nor any extra guidance inputs. In both training and inference, GPVA uses the robot state at $t_0$ to compute the attention bias for action queries in the target chunk.

\subsection{Supplementary Explanations of CAPA}
\label{sec:supp_capa}
\label{sec:supp_capa_details}


\paragraph{The surrogate-gradient design.}
CAPA selects a retained subset of video tokens for each frame before the transformer backbone. Since the selection is discrete, the task loss is non-differentiable through selected indices. We therefore adopt the surrogate in Eq.~\eqref{eq:capa-surrogate} in the main paper to provide an available training signal meanwhile keeping the forward computation unchanged.

For each frame, let $r_i$ denote the retention score after sigmoid of the $i$-th token. 
For selected tokens, the surrogate weight is $g_i=r_i$.
The correction term in Eq.~\eqref{eq:capa-surrogate},
$\left(g_i-\operatorname{sg}(g_i)\right)
\operatorname{sg}(\hat h^v_{\ell,i}-h^v_{\ell,i})$,
is zero in the forward propagation, hence the forward computation remains unchanged. In the backward propagation, the correction term provides a gradient to $g_i$ based on the detached $\hat h_{\ell,i}^v-h_{\ell,i}^v$. Hence an approximate gradient is supplied to the scoring network.
The transformer backbone still receives the standard gradient through $\hat h_{\ell,i}^v$. As described in Sec.~\ref{sec:capa} in the main paper, retained video tokens are used across all layers in the transformer backbone. Accordingly, gradient contributions from all layers are accumulated on $g_i$.
The chain rule gives:
\begin{equation}
    \frac{\partial\mathcal L}{\partial g_i}
    =\sum_{\ell}
    \left(
        \frac{\partial\mathcal L}{\partial h_{\ell+1,i}^{v}}
    \right)^{\!\top}
    \operatorname{sg}(\hat h_{\ell,i}^v-h_{\ell,i}^v).
    \label{eq:supp_capa_accumulated_gradient}
\end{equation}


\paragraph{Essential impacts of CAPA.}
The scoring network is trained jointly with the transformer backbone using the
video-action loss computed over all action tokens and retained video tokens learned from observed
scene contents and temporal changes.
Under the restricted budget, by preferentially retaining tokens from spatial positions that are informative for future video prediction and action generation, CAPA enables task-relevant attention allocation, meanwhile reducing computational burden on redundant
backgrounds. Overall, CAPA enables AG-WAM to predict future videos and actions from a compact set of informative video tokens.

\subsection{The Definition of Visual Attention Matching Rate}
\label{sec:supp_matching_rate}

As shown in Fig.~\ref{fig:Fig6} in the main paper, to quantitatively evaluate whether action queries attend to interaction-relevant visual regions, we compute the Visual Attention Matching Rate (\%) over trajectories as follows:
\begin{enumerate}[leftmargin=*, itemsep=5pt, topsep=5pt]
    \item \textbf{Ground-truth grid cell}: For each trajectory, we first extract per-frame ground-truth guidance points $\{p_t\}$ based on Alg.~\ref{alg:guidance_extraction}. At frame $t$, $p_t$ is projected onto a certain camera view, visualized as the red circle in Fig.~\ref{fig:Fig6}. 
    Consistent with Sec.~\ref{sec:capa} in the main paper, the image is partitioned into an $H_g \times W_g$  grid, and the grid cell containing the projected point is designated as the ground-truth cell $(r_{\mathrm{gt}}, c_{\mathrm{gt}})$ highlighted by the red box in Fig.~\ref{fig:Fig6}. 
    \item \textbf{Prediction grid cell}: At the same timestep $t$, we extract the predicted action-to-visual attention weights over the $H_g \times W_g$ grid (averaged over action tokens and attention heads in the GPVA-equipped layers), and locate the grid cell with the maximum attention response highlighted by the green box in Fig.~\ref{fig:Fig6}:
    \begin{equation}
        (r_{\mathrm{pred}}, c_{\mathrm{pred}}) = \operatorname*{arg\,max}_{(r,c) \in \{1,\dots,H_g\} \times \{1,\dots,W_g\}} A_t(r, c),
    \end{equation}
    \item \textbf{$\mathbf{8}$-neighborhood matching criterion}: A frame is counted as the matched frame only if the prediction  cell lies within the $8$-connected neighborhood of the ground-truth cell, \emph{i.e.} under the Chebyshev grid distance:
    \begin{equation}
        \max\bigl(|r_{\mathrm{pred}} - r_{\mathrm{gt}}|,\, |c_{\mathrm{pred}} - c_{\mathrm{gt}}|\bigr) \le 1,
    \end{equation}
\end{enumerate}
The Visual Attention Matching Rate reported in Fig.~\ref{fig:Fig6} is defined as the percentage of matched frames over all valid frames across all  trajectories on each dataset. 

\begin{table}[t]
\centering
\caption{Performance stability of AG-WAM on LIBERO-X over $3$ independent evaluation runs.}
\vspace{-0.3cm}
\label{tab:stability_liberox}
\footnotesize
\renewcommand{\arraystretch}{1.15}
\setlength{\tabcolsep}{4pt}
\resizebox{\linewidth}{!}{%
\begin{tabular}{lcccccc}
\toprule
Metric & L1 & L2 & L3 & L4 & L5 & Average \\
\midrule
Mean & 79.2\% & 69.0\% & 44.3\% & 33.5\% & 24.8\% & 50.2\% \\
Std  & 0.2\% & 0.1\% & 0.2\% & 0.2\% & 0.3\% & 0.1\% \\
\bottomrule
\end{tabular}%
}
\end{table}

\begin{table}[t]
\centering
\caption{Performance stability of AG-WAM on LIBERO-Plus over $3$ independent evaluation runs.}
\vspace{-0.3cm}
\label{tab:stability_liberoplus}
\small
\renewcommand{\arraystretch}{1.15}
\setlength{\tabcolsep}{12pt}
\begin{tabular}{lcc}
\toprule
Dimension & Mean & Std \\
\midrule
Camera & 98.4\% & 0.1\% \\
Robot & 94.5\% & 0.2\% \\
Language & 90.1\% & 0.2\% \\
Light & 98.1\% & 0.2\% \\
Background & 99.0\% & 0.3\% \\
Layout & 88.9\% & 0.4\% \\
Noise & 98.4\% & 0.3\% \\
Overall & 95.1\% & 0.1\% \\
\bottomrule
\end{tabular}
\end{table}

\begin{table}[t]
\centering
\caption{Performance stability of AG-WAM on RoboTwin-2.0 over $3$ independent evaluation runs.}
\vspace{-0.3cm}
\label{tab:stability_robotwin}
\small
\renewcommand{\arraystretch}{1.15}
\setlength{\tabcolsep}{12pt}
\begin{tabular}{lcc}
\toprule
Setting & Mean & Std \\
\midrule
Clean & 97.06\% & 0.12\% \\
Random & 97.22\% & 0.24\% \\
Average & 97.14\% & 0.13\%\\
\bottomrule
\end{tabular}
\end{table}

\section{Additional Experiments}

\subsection{Performance Stability on Three Benchmarks}
\label{sec:stable}
To verify the stability and reproducibility of our reported state-of-the-art performance on three public benchmarks, we conduct three independent evaluation runs on each benchmark, and report both the mean and standard deviation (std) of the success rate (\%):
\begin{itemize}[leftmargin=*, itemsep=3pt, topsep=3pt]
    \item \textbf{LIBERO-X}: spans five evaluation levels: L1$\sim$L5, $600$ tasks $\times$ $10$ trajectories each for L1$\sim$L3, $826$ tasks $\times$ $10$ trajectories for L4, and $826$ tasks $\times$ $10$ trajectories for L5, totaling $34{,}520$ evaluation trajectories per run.
    \item \textbf{LIBERO-Plus}: spans $7$ perturbation dimensions: Camera ($1{,}599$), Robot ($1{,}550$), Language ($1{,}537$), Light ($1{,}142$), Background ($1{,}076$), Noise ($1{,}601$), and Layout ($1{,}525$), totaling $10{,}030$ evaluation trajectories per run. 
    \item \textbf{RoboTwin-2.0}: evaluates $50$ bimanual manipulation tasks with $100$ trajectories per task under both clean and randomized settings, totaling $10{,}000$ evaluation trajectories per run. 
\end{itemize}
Tab.~\ref{tab:stability_liberox}, Tab.~\ref{tab:stability_liberoplus}, and Tab.~\ref{tab:stability_robotwin} summarize the performance stability (mean and standard deviation) of AG-WAM on all three benchmarks, matching the evaluation protocols of Tab.~\ref{tab:liberox_main}, Tab.~\ref{tab:liberoplus_main}, and Tab.~\ref{tab:robotwin_main} in the main paper respectively.

\subsection{Ablation Studies on LIBERO-Plus and RoboTwin-2.0}
\label{sec:supp_more_ablations}

To further verify that the component-wise benefits of our attention guidance modules achieved on LIBERO-X (Tab.~\ref{tab:liberox_ablation} in the main paper) can generalize across perturbation robustness and bimanual manipulation, we conduct further ablation studies on LIBERO-Plus  and RoboTwin-2.0.

\begin{table}[t]
\centering
\caption{Ablation studies on LIBERO-Plus. We report the overall success rate (\%) across seven perturbation dimensions.}
\vspace{-0.3cm}
\label{tab:liberoplus_ablation}
\small
\renewcommand{\arraystretch}{1.15}
\setlength{\tabcolsep}{12pt}
\begin{tabular}{lc}
\toprule
Method & Overall \\
\midrule
Baseline(BS) & 90.4\% \\
BS + CAPA & 90.7\% \\
BS + SCTA & 93.0\% \\
BS + GPVA & 94.1\% \\
BS + SCTA + GPVA & 94.9\% \\
\textbf{AG-WAM} & \textbf{95.1\%} \\
\bottomrule
\end{tabular}
\end{table}

\begin{table}[t]
\centering
\caption{Ablation studies on RoboTwin-2.0.}
\vspace{-0.3cm}
\label{tab:robotwin_ablation}
\scriptsize
\renewcommand{\arraystretch}{1.15}
\setlength{\tabcolsep}{10pt}
\begin{tabular}{lccc}
\toprule
Method & Clean & Random & Average \\
\midrule
Baseline(BS) & 94.12\% & 92.60\% & 93.36\% \\
BS + CAPA & 94.34\% & 92.72\% & 93.53\% \\
BS + SCTA & 95.76\% & 94.68\% & 95.22\% \\
BS + GPVA & 96.38\% & 96.26\% & 96.32\% \\
BS + SCTA + GPVA & 96.98\% & 97.04\% & 97.01\% \\
\textbf{AG-WAM} & \textbf{97.06\%} & \textbf{97.22\%} & \textbf{97.14\%} \\
\bottomrule
\end{tabular}
\end{table}

\begin{table*}[t]
\centering
\caption{Intra-module ablation studies of CAPA on LIBERO-X. The average success rate (\%) over level L1$\sim$L5 is adopted for evaluation.}
\vspace{-0.3cm}
\label{tab:capa_intra_ablation}
\small
\renewcommand{\arraystretch}{1.15}
\setlength{\tabcolsep}{8pt}
\begin{tabular}{lcccc}
\toprule
Variant & Retention Ratio & Predictive Ratio & Random Ratio & Average \\
\midrule
Baseline(BS) & 100\% & -- & -- & 44.0\% \\
BS + Random Sampling & 60\% & 0\% & 60\% & 40.5\% \\
BS + CAPA (w/o random) & 60\% & 60\% & 0\% & 44.1\% \\
BS + CAPA (random 5\%) & 60\% & 55\% & 5\% & 44.1\% \\
\textbf{BS + CAPA (random 10\%)} & 60\% & 50\% & 10\% & \textbf{44.3\%} \\
BS + CAPA (random 20\%) & 60\% & 40\% & 20\% & 43.6\% \\
\bottomrule
\end{tabular}
\end{table*}

\begin{table*}[t]
\centering
\caption{Intra-module ablation studies of GPVA on LIBERO-X. The average success rate (\%) over level L1$\sim$L5 is adopted for evaluation.}
\vspace{-0.3cm}
\label{tab:gpva_intra_ablation}
\small
\renewcommand{\arraystretch}{1.15}
\setlength{\tabcolsep}{6pt}
\resizebox{\textwidth}{!}{%
\begin{tabular}{lcccccc}
\toprule
Variant & 2D $\mathcal{L}_{\mathrm{MSE}}$ & 2D $\mathcal{L}_{\mathrm{JS}}$ & 3D $\Delta\hat{p}_t$ & 3D $\sigma_t$ & Bias Injection & Average \\
\midrule
Baseline(BS) & -- & -- & -- & -- & -- & 44.0\% \\
BS + Aux.\ Loss & \checkmark & \checkmark & \checkmark & \checkmark & $\times$ & 44.7\% \\
BS + 2D $\mathcal{L}_{\mathrm{MSE}}$ & \checkmark & $\times$ & $\times$ & $\times$ & $b_t^{\mathrm{2D}}$ & 46.5\% \\
BS + 2D Full & \checkmark & \checkmark & $\times$ & $\times$ & $b_t^{\mathrm{2D}}$ & 47.0\% \\
BS + 3D $\Delta\hat{p}_t$ & $\times$ & $\times$ & \checkmark & $\times$ & $b_t^{\mathrm{3D}}$ & 45.7\% \\
BS + 3D Full & $\times$ & $\times$ & \checkmark & \checkmark & $b_t^{\mathrm{3D}}$ & 46.3\% \\
BS + 2D Full + 3D Full, static & \checkmark & \checkmark & \checkmark & \checkmark & $0.5\,b_t^{\mathrm{2D}} + 0.5\,b_t^{\mathrm{3D}}$ & 47.7\% \\
\textbf{BS + GPVA} & \checkmark & \checkmark & \checkmark & \checkmark & $\beta_t b_t^{\mathrm{2D}} + (1-\beta_t) b_t^{\mathrm{3D}}$ & \textbf{48.4\%} \\
\bottomrule
\end{tabular}%
}
\end{table*}

\begin{table*}[t]
\centering
\caption{Intra-module ablation studies of SCTA on LIBERO-X. The average success rate (\%) over level L1$\sim$L5 is adopted for evaluation. }
\vspace{-0.3cm}
\label{tab:scta_intra_ablation}
\small
\renewcommand{\arraystretch}{1.15}
\setlength{\tabcolsep}{8pt}
\begin{tabular}{lcccc}
\toprule
Variant & Modulation & $\mathcal{L}_{\mathrm{ent}}$ & $[\tau_{\mathrm{low}}, \tau_{\mathrm{high}}]$ & Average \\
\midrule
Baseline(BS) & -- & -- & -- & 44.0\% \\
BS + global-context-only & \checkmark & $\times$ & -- & 44.6\%  \\
BS + Modulation & \checkmark & $\times$ & -- & 45.7\% \\
BS + SCTA (dispersive) & \checkmark & \checkmark & $[0.6, 0.9]$ & 45.2\% \\
BS + SCTA (concentrated) & \checkmark & \checkmark & $[0.05, 0.2]$ & 45.9\% \\
\textbf{BS + SCTA (intermediate)} & \checkmark & \checkmark & $[0.2, 0.6]$ & \textbf{46.4\%} \\
\bottomrule
\end{tabular}
\end{table*}

\paragraph{Ablation studies on LIBERO-Plus.}
Tab.~\ref{tab:liberoplus_ablation} reports the overall success rate (\%) across seven perturbation dimensions. Consistent with the results on LIBERO-X, applying CAPA with $40\%$ of video tokens discarded slightly improves by $0.3\%$ over the baseline model meanwhile reducing computational overhead. Integrating SCTA and GPVA  individually improves overall success rates by $2.6\%$  and $3.7\%$ respectively. Integrating all three attention guidance modules, AG-WAM achieves state-of-the-art performance on LIBERO-Plus, with the overall success rate improvement of $4.7\%$ over the baseline model.

\paragraph{Ablation studies on RoboTwin-2.0.}
Consistent with the results on LIBERO-X, as shown in Tab.~\ref{tab:robotwin_ablation}, applying CAPA with $40\%$ of video tokens discarded slightly improves by $0.17\%$ over the baseline model meanwhile reducing computational overhead. Integrating SCTA and GPVA  individually improves average success rates by $1.86\%$  and $2.96\%$ respectively. Integrating all three attention guidance modules, AG-WAM achieves state-of-the-art performance on RoboTwin-2.0, with the average success rate improvement of $3.78\%$ over the baseline model.

\subsection{Intra-Module Ablation Studies of CAPA}
\label{sec:supp_capa_ablation}

To further evaluate the internal design in CAPA, we conduct detailed intra-module ablation studies on LIBERO-X built upon the baseline model from Tab.~\ref{tab:liberox_ablation} in the main paper. As shown in Tab.~\ref{tab:capa_intra_ablation}, retaining $60\%$ of video tokens all from random selection severely degrades the performance by $3.5\%$ compared with the baseline model, whereas retaining $60\%$ of video tokens all from predictive selection through CAPA keeps the performance stable. In order to additionally introduce robustness into token selection, three variants with random selection ratios of $5\%$, $10\%$, and $20\%$ are evaluated respectively, in which the random selection ratio of $10\%$ achieves the best performance with an improvement of $0.3\%$, meanwhile reducing computational overhead as shown in Tab.~\ref{tab:liberox_efficiency} in the main paper.

\subsection{Intra-Module Ablation Studies of GPVA}
\label{sec:supp_gpva_ablation}

To further evaluate the internal design in GPVA, we conduct detailed intra-module ablation studies on LIBERO-X built upon the baseline model from Tab.~\ref{tab:liberox_ablation} in the main paper. 
Tab.~\ref{tab:gpva_intra_ablation} examines the contribution of the 2D heatmap supervision ($\mathcal{L}_{\mathrm{MSE}}$), the Jensen-Shannon divergence loss ($\mathcal{L}_{\mathrm{JS}}$), the 3D offset regression ($\Delta\hat{p}_t$), the 3D spatial uncertainty estimation ($\sigma_t$), and the mechanism of attention-logit bias injection. The experimental results clearly demonstrate the effectiveness of our design:
\begin{enumerate}[leftmargin=*, itemsep=5pt, topsep=5pt]
    \item \textbf{Explicit bias injection \emph{vs} auxiliary supervision}: Adding auxiliary supervision alone yields only a slight gain over the baseline model by $0.7\%$, whereas enabling the explicit attention-logit bias injection achieves an improvement of $4.4\%$, confirming that the primary improvement derives from the explicit attention modulation rather than the multi-task representation learning alone.
    \item \textbf{Complementarity between 2D and 3D branches}: Integrating the 2D branch and the 3D branch individually outperforms the baseline model by $3.0\%$ and $2.3\%$ respectively. Combining both branches via the adaptive fusion yields a notable improvement of $4.4\%$, clearly demonstrating the complementarity between 2D appearance priors and 3D geometric priors.
    \item \textbf{Distributional regularization and spatial uncertainty}: Incorporating $\mathcal{L}_{\mathrm{JS}}$ into the 2D branch  improves the success rate by $0.5\%$, validating the  benefit of suppressing low-amplitude spurious peaks on distractors. Similarly, integrating the 3D spatial uncertainty estimation $\sigma_t$ into the 3D branch  boosts the success rate by $0.6\%$, revealing that confidence-aware 3D querying produces more reliable spatial reference distributions.
    \item \textbf{Adaptive \emph{vs} static fusion}: Replacing the simple static averaging combination with the learned adaptive fusion brings an additional improvement of $0.7\%$, validating the benefit of dynamically balancing between 2D and 3D attention biases across frames and viewpoints.
\end{enumerate}

\subsection{Intra-Module Ablation Studies of SCTA}
\label{sec:supp_scta_ablation}

To further evaluate the internal design in SCTA, we conduct detailed intra-module ablation studies on LIBERO-X built upon the baseline model from Tab.~\ref{tab:liberox_ablation} in the main paper. Tab.~\ref{tab:scta_intra_ablation} examines the contribution of the instruction-guided action modulation, the interval-constrained entropy loss ($\mathcal{L}_{\mathrm{ent}}$), and the sensitivity to different normalized entropy bounds ($[\tau_{\mathrm{low}}, \tau_{\mathrm{high}}]$).
The experimental results clearly demonstrate the effectiveness of our design:
\begin{enumerate}[leftmargin=*, itemsep=5pt, topsep=5pt] 
    \item \textbf{Global-context-only modulation}: We introduce a global-context-only variant that removes SCTA's multi-head attention over $E_{text}$, and only uses the global task summary obtained by averaging $N_{text}$ instruction token embeddings in $E_{text}$, to determine the ``scale and shift'' in the FiLM. The FiLM is applied only on the noisy action stream. As can be seen, this variant is inferior to our proposed instruction-guided action modulation, verifying the benefit of incorporating state-conditioned text semantics in addition to the global task summary.
    \item \textbf{Instruction-guided action modulation}: Integrating the instruction-guided action modulation improves the success rate over the baseline model by $1.7\%$, confirming the benefit of conditioning action features based on the evolving execution state.
    \item \textbf{Interval-constrained entropy loss}: Enforcing an overly dispersive interval ($[0.6, 0.9]$) degrades the performance by $0.5\%$, as it encourages flat attention weights across irrelevant text segments thus diluting the gain of SCTA. Conversely, an overly concentrated interval ($[0.05, 0.2]$) yields only a limited improvement of $0.2\%$, since forcing attention onto very few tokens risks the loss of context. By contrast, the intermediate interval ($[0.2, 0.6]$) strikes an appropriate balance, boosting the success rate by $0.7\%$, demonstrating the effectiveness of the interval-constrained entropy loss.
\end{enumerate}

\subsection{Supplementary Experimental Settings}
\label{sec:supp_exp_setting}

Similar to \cite{li2026lingbotva,xiang2026motubrain}, 
we adopt the multi-stage training setup. Starting from the publicly available Internet-video pretrained Wan2.2-5B \cite{wan2025wan} 
weights, we first pretrain our basic weights using an embodied training data corpus by aggregating four existing public robot manipulation datasets: RoboCoin~\cite{wu2025robocoin}, Agibot~\cite{world2025agibot}, RoboMind~\cite{wu2024robomind}, and InternData-A1~\cite{tian2026interndata}. 
We train exclusively on publicly available data, without using any extra private embodied data. Then, based on the same pretrained basic weights, we perform supervised finetuning (SFT) on each target dataset (LIBERO-X, LIBERO-Plus, and RoboTwin-2.0) separately 
for all models except AG-WAM\_zeroshot: our baseline model, models in ablation studies, and our AG-WAM.
For SFT on LIBERO-X, we use only its publicly released training data including $2,520$ trajectories. 
For SFT on LIBERO-Plus, we use only its publicly released training data with perturbation including a total of $15,874$ trajectories. 
For SFT on RoboTwin-2.0, we use only its publicly released clean and randomized training data, including $2,500$ trajectories collected in clean scenes ($50$ per task) and $25,000$ trajectories collected in randomized scenes ($500$ per task).

The overall training loss is formulated as:
\begin{equation}
\label{eq:loss_overall}
\begin{aligned}
\mathcal{L}_{\mathrm{total}}
= \mathcal{L}_{\mathrm{video}} + \lambda_\mathrm{a}{\mathcal{L}_{\mathrm{action}}} + \lambda_\mathrm{GPVA}{\mathcal{L}_{\mathrm{GPVA}}} + \lambda_\mathrm{ent}{\mathcal{L}_{\mathrm{ent}}}
\end{aligned}
\end{equation}
where $\lambda_\mathrm{a}$, $\lambda_\mathrm{GPVA}$, and $\lambda_\mathrm{ent}$ are set to $1.0$, $1.0$, and $0.2$ respectively. At inference, we use an Euler ODE solver with the video CFG scale $5.0$ and the action CFG scale $1.0$.

\end{document}